%% file: mainv5.tex
\documentclass[11pt,a4paper]{article}
\usepackage[hyperref]{main}
\usepackage{times}
\usepackage{latexsym}
\usepackage{alltt}
\usepackage{amsmath}

\usepackage[utf8]{inputenc}
\usepackage{pgfplots}
\usepackage{subfig}

\usepackage{tikz}
\usetikzlibrary{calc}
\usetikzlibrary{positioning}
\usetikzlibrary{shapes.multipart}
\usepackage{relsize}

\pgfplotsset{
every axis/.append style={
ylabel shift=-1pt,
xlabel shift=-1pt,
xlabel near ticks,
ylabel near ticks,
font=\small
},
compat=1.14
}
\newlength\figureheight
\newlength\figurewidth

\usepackage{url}

\aclfinalcopy % Uncomment this line for the final submission

\title{Evaluating the Ability of LSTMs to Learn Context-Free Grammars}

\author{Luzi Sennhauser \\
  Federal Institute of Technology\\
  Zurich, Switzerland \\
  Massachusetts Institute of Technology\\
  Cambridge, MA, USA \\
  {\tt luzis@student.ethz.ch} \\\And
  Robert C. Berwick\\
  LIDS, Room 32-D728\\ 
  Massachusetts Institute of Technology \\
  Cambridge, MA, USA \\
  {\tt berwick@csail.mit.edu}
}

\date{}

\begin{document}
\maketitle
\begin{abstract}

While long short-term memory (LSTM) neural net architectures are designed to capture sequence information, human language is generally composed of hierarchical structures. This raises the question as to whether LSTMs can learn hierarchical structures. We explore this question with a well-formed bracket prediction task using two types of brackets modeled by an LSTM.

Demonstrating that such a system is learnable by an LSTM is the first step in demonstrating that the entire class of CFLs is also learnable. We observe that the model requires exponential memory in terms of the number of characters and embedded depth, where a sub-linear memory should suffice.

Still, the model does more than memorize the training input. It learns how to distinguish between relevant and irrelevant information. On the other hand, we also observe that the model does not generalize well.

We conclude that LSTMs do not learn the relevant underlying context-free rules, suggesting the good overall performance is attained rather by an efficient way of evaluating nuisance variables. LSTMs are a way to quickly reach good results for many natural language tasks, but to understand and generate natural language one has to investigate other concepts that can make more direct use of natural language's structural nature.
\end{abstract}

\section{Introduction}

Composing hierarchical structure for natural language is an extremely powerful tool for human language generation.  These structures are of great importance in order to extract semantic interpretation \cite{berwick2016only} and enable us to produce a vast repertoire of sentences via a very small set of rules. Having acquired such a set of rules, it is easy to construct new structures without having previously seen similar examples.

For purposes of external communication, the syntactic structures generated by grammars must be ``flattened'' or linearized into a sequential output form (e.g. written, signed, or spoken). When reading such a (linearized) text, hearing a spoken sentence or observing a signed language, the structure has to be recovered implicitly to recover the original meaning (i.e., parsing).

In this study, we investigate whether Long Short-Term Memory (LSTM) models \cite{hochreiter1997long} possess this same ability as humans do: inferring rule-based structure from a linear representation. \citeauthor{everaert2015structures} \shortcite{everaert2015structures} show clearly that there are phenomena in human language that can only be understood by taking the underlying hierarchical structure into account. For neural networks to do the same, it is therefore essential to acquire the underlying structure of sentences.

Recurrent neural networks are often used for tasks like language modeling \cite{mikolov2010recurrent, sundermeyer2012lstm}, parsing \cite{vinyals2015grammar, kiperwasser2016simple, dyer2016recurrent}, machine translation \cite{bahdanau2014neural}, and morphological compositions \cite{kim2016character}. LSTMs are inherently sequential models. Since the hierarchical structures appearing in natural language often correlate with sequential statistical features, it can be difficult to evaluate whether an LSTM learns the underlying rules of the sentence's syntax or alternatively simply learns sequential statistical correlations.  In this paper we carry out experiments to determine this.

We set up our experiments by posing the LSTM with a bracket completion problem having two possible bracket types, a so-called Dyck Language. A model which recognizes this language has to infer rules of the underlying structure. Furthermore, a system that can solve this task is able to recognize every context-free grammar (see section \ref{sec:corpus} regarding Dyck Languages via the Chomsky-Sch\"utzenberger theorem for why this is so).

By analyzing the intermediate states of the corresponding LSTM networks, observing generalization behaviours, and evaluating the memory demands of the model we investigate whether LSTMs acquire rules as opposed to statistical regularities.

\section{Related work}
It has been shown that LSTMs are able to count and partly acquire for context-free languages like $a^nb^n$ and simple context-sensitive languages \cite{gers2001lstm, rodriguez2001simple}. We note that in contrast to the language we investigate here, $a^nb^n$ may be considered the ``simplest'' context-free language, since it can be generated by a grammar with just one transition.

The question as to whether LSTMs can infer rules on a natural language corpus, e.g., for subject-verb agreement, was initially explored by others such as \cite{linzen2016assessing}. \citeauthor{liska2018memorize} \shortcite{liska2018memorize} investigated the memorization vs. generalization issue for LSTMs for function composition: they showed that if an LSTM learns the mapping from a string-set A to B and from B to C, then the direct mapping from A to C can partly be learned. We use the same method and model for a different task -- instead of function composition we evaluate it for bracket matching.

Since most of the time it is challenging to determine what is actually going on with respect to the neural network's internal state, several attempts have been made to visualize a neural network's intermediate states with the goal of making them interpretable \cite{rauber2017visualizing, karpathy2015visualizing, krakovna2016increasing}. For several simple copy and palindrome language tasks, it has been shown that RNNs learn a fractal encoding similar to a binary expansion of the input \cite{tabor2000fractal, gruening2006stack, kirov2012processing}. With the same objective we use another, recently introduced method to investigate the internal states.

While here we investigate the ability of how well structural information can be stored in originally sequential models, other approaches are currently being taken to move from sequential models to structural ones, e.g. to hardwire structural properties into the model's architecture \cite{tai2015improved, kiperwasser2016simple, joulin2015inferring}; to make a larger external memory available to the network \cite{graves2014neural, sukhbaatar2015end}; or to make the network architecture dynamic \cite{moshe2017deep}.

Finally, we note that thanks to careful reviewing, we were made aware of \citeauthor{bernardy2018can}'s work \shortcite{bernardy2018can}, that addresses essentially the same task as the one we tackle: He investigated also the generalization behaviour of LSTMs for a Dyck-language corpus with several bracket types. He investigated generalization for sentences by concatenating several training sentences; or embedding training sentences in a centrally embedded bracket string. In contrast, we evaluate generalization by training sentences on a certain feature (number of characters, embedded depth) and testing the resulting model on the out-of-sample sentences. By this method, we strive to reduce the probability of similar sub-strings in the training versus the test set.

\section{Corpus}
\label{sec:corpus}
When dealing with natural language, there are many side effects or nuisance variables  -- e.g. words occurring more often in certain correlative contexts or clusters than others.  These can influence any classification and experimental result. To minimize such effects, we conducted all experiments on artificial corpora.

The Chomsky-Sch\"utzenberger theorem \cite{chomsky1963algebraic, autebert1997context} about representing context-free language (CFL) states the following: ``For  each context-free language $L$, there is a positive integer $n$, a regular language $R$, and a homomorphism $h$ such that $L=h(D_n \cup R)$.'' where $D_n$ is a Dyck language with $n$ different bracket pairs. As described by \citeauthor{forisek2018what} \shortcite{forisek2018what}, it follows that the Dyck language $D_2$ essentially covers the entire class of CFLs. Every model which recognizes or generates well-formed Dyck words with two types of brackets should be powerful enough to handle any CFL when intersected with a relabeling (homomorphism of a constructed regular language).

The synthetic corpus we use consists of such a Dyck language with two types of brackets (\verb|[]| and \verb|{}|). Sentences are generated according to the following grammar:

\begin{samepage}
\begin{alltt}
    S  -> S1 S | S1
    S1 -> B | T
    B  -> [ S ] | \{ S \}
    T  -> [ ] | \{ \}
\end{alltt}
\end{samepage}

The probabilities of the rules are defined in a way that the entropy -- in terms of the number of characters between an opening and its corresponding closing bracket and the depth of embedding at which a bracket appears -- is larger than if the rules had all equal probabilities. Formally, the branching probability $P_b = P[\texttt{S1 -> B}]$ and the concatenation probability $P_c = P[\texttt{S -> S1 S}]$ are defined as follows:

\begin{equation}
\begin{split}
    s(l) &= min(1, -3 \cdot \frac{l}{n}+3) \\
    P_b &= r_b \cdot s(l) \quad \textrm{where} \quad r_b\sim\mathcal{U}(0.4,0.8)\\
    P_c &= r_c \cdot s(l) \quad \textrm{where} \quad r_c\sim\mathcal{U}(0.4,0.8)
\end{split}
\end{equation}

where $r_b$ and $r_c$ are sampled once per sentence and $l$ is the number of already generated characters in the sentence.
All 1M generated sentences have a length $n$ of 100 characters.

In this paper, we check whether an LSTM can be trained to recognize this grammar.

\setlength\figureheight{4cm}
\setlength\figurewidth{\linewidth}
\begin{figure}[ht]
    \input{figures/corpus_distance_frequencies.tex}\\%
    \input{figures/corpus_depth_frequencies.tex}%
    \caption{Corpus frequencies}%
    \label{fig:corpus_frequencies}%
\end{figure}
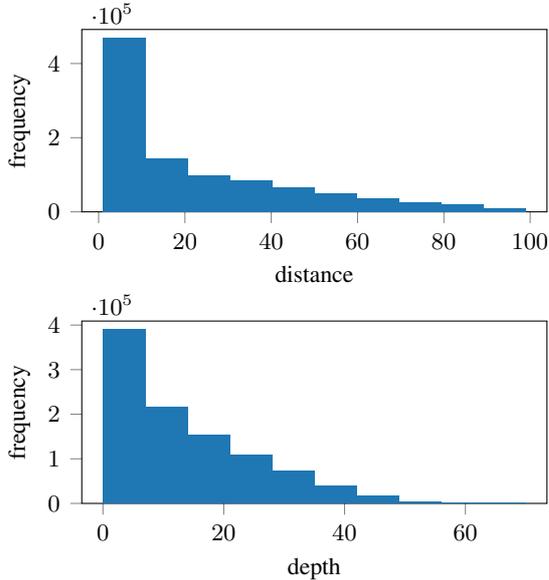

\section{Model}

To check if we can train a neural network to accept the language generated by the grammar above, an LSTM is used.

\subsection{Long Short-Term Memory}

Long-Short-Term-Memory networks (LSTM) \cite{hochreiter1997long} are a variant of recurrent neural networks (RNNs). Both of them possess a memory state that is updated in the process of reading a time series. Many RNNs suffer from the problem of vanishing gradients \cite{hochreiter1997long}: The recurrent activation functions of RNNs are often set to be $tanh$ or the sigmoid function. Since their gradients are most of the times smaller than 1 (for $tanh$ it is upper bounded by $1$, and for the sigmoid function even by $0.25$), the gradient cannot be conserved during extense backpropagation and approaches 0. LSTMs deal with this issue by containing three multiplicative gates controlling what proportion of the input to pass to the memory cell (input gate), what proportion of the previous memory cell information to discard (forget gate) and what proportion of the memory cell to output (output gate). In the recurrency of the LSTM the activation function is the identity function, which has gradient $1.0$. This means that if the forget gate is open, the gradient is fully passed on to previous time steps, and long term dependencies can be learned.

The LSTM reads each input $x_i$ consecutively and updates its memory state $c_i$ accordingly. After each step, an output $h_i$ is generated based on the updated memory state. More specifically, the LSTM solves the following equations in a forward pass:
\begin{equation}
\begin{split}
    \mathbf{i}_t &= \sigma(\mathbf{W}_{ix}\mathbf{x}_t+\mathbf{W}_{ih}\mathbf{h}_{t-1}+\mathbf{b}_i) \\
    \mathbf{f}_t &= \sigma(\mathbf{W}_{fx}\mathbf{x}_t+\mathbf{W}_{fh}\mathbf{h}_{t-1}+\mathbf{b}_f) \\
    \mathbf{c}_t &= \mathbf{f}_t \odot \mathbf{c}_{t-1}\\
    & \qquad +\mathbf{i}_t \odot tanh(\mathbf{W}_{cx}\mathbf{x}_t + \mathbf{W}_{ch}\mathbf{h}_{t-1}+\mathbf{b}_c) \\
    \mathbf{o}_t &= \sigma(\mathbf{W}_{ox}\mathbf{x}_t + \mathbf{W}_{oh}\mathbf{h}_{t-1} + \mathbf{b}_o)\\
    \mathbf{h}_t &= \mathbf{o}_t \odot tanh(\mathbf{c}_t) 
    \label{eq:lstm_equations}
\end{split}
\end{equation}

\begin{figure}[ht]
    \centering
    \input{figures/lstm_cell.tex}
    \caption{schematic model of an LSTM-cell. $\bigodot$ stands for element-wise multiplication and $\bigoplus$ for vector addition.}
    \label{fig:lstm_cell}

\end{figure}
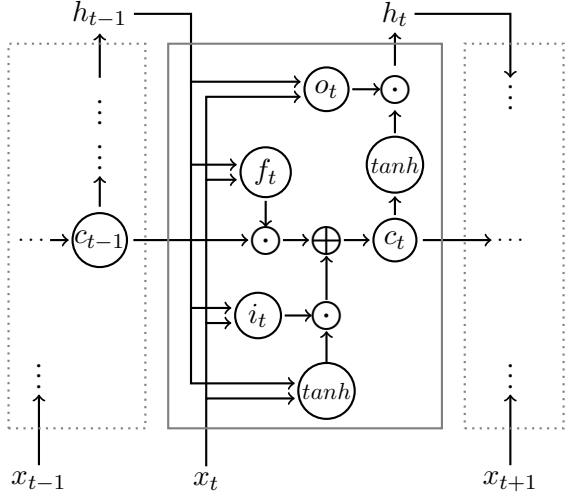
\subsection{Basic Model}
\label{subsec:basic_model}
Now let us turn to the details of the model implementation.  We begin with the basic formulation.

Let $B_{open}$ and $B_{close}$ be the sets of opening and closing brackets and $B = B_{open} \cup B_{close}$ the set of all brackets. Given the beginning of a sentence $w_1, w_2, ..., w_{k-1}, w_k$ with $w_1, ..., w_{k-1} \in B$ and $w_k \in B_{close}$, the LSTMs tries approximate the function:
\begin{align*}
  F \colon B^{k-1} &\to B_{close}\\
  w_1, w_2, ..., w_{k-1} &\mapsto w_k.
  \label{eq:brackets_task_definition}
\end{align*}
The substring (clause) between the corresponding opening bracket of $w_i$ and $w_i$ will be referred to as the \emph{relevant clause} in the remainder of this paper. Likewise, by \emph{distance} we denote the number of characters of the relevant clause. Note that this distance is always a multiple of 2, since the relevant clause is well-balanced. The \emph{depth} at a certain position $i$ is the number of unclosed brackets in the first $i$ characters. The \emph{embedded depth} of a sentence is the maximum \emph{depth} when processing the relevant clause.

%Note that since a bracket pair consists of 2 characters, distances are always even. We denote \emph{odd distances} as ${2,6,10,14,...}$ and \emph{even distances} as ${4,8,12,16,...}$.

To read the input characters, an embedding layer with 5 output dimensions precedes the LSTM. Together they build the encoder, which will read the input sequentially. The decoder, mapping the internal representation to a probability of predicting \verb|}| or \verb|]| is a dense layer with one output variable.

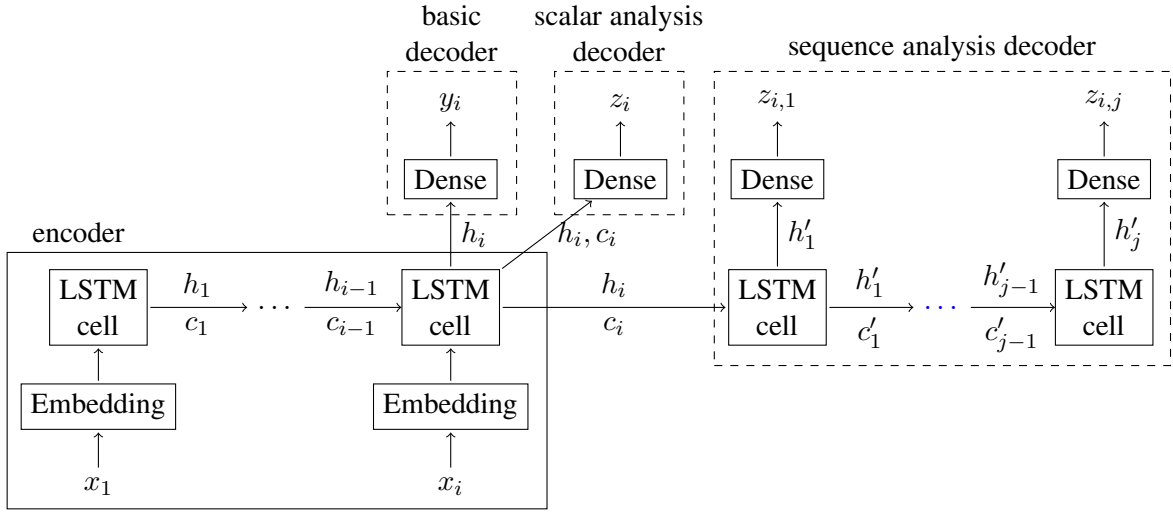
\begin{figure*}[ht]
    \centering
    \input{figures/lstm_architecture.tex}
    \caption{Network architecture of the model. The basic end-to-end model consists of the encoder and the basic decoder. The analysis model fixes the weights for the encoder and uses the scalar (if $z_i$ is a scalar) or sequence analysis decoder (if $z_i$ is a sequence).}
    \label{fig:lstm_architecture}
\end{figure*}

We have compared different initialization methods. It turns out that the initialization of the model is crucial to avoid bad local minima. The following initialization method results in consistently good solutions: To initialize the weights, the model is trained with sentences of length 50 and only afterwards on the actual corpus with sentence length 100.

For backpropagation, the Adam \cite{kingma2014adam} optimizer was used. Furthermore, to ensure faster and more consistent convergence, at the beginning of the training, the batch size is gradually increased, which has a similar effect as reducing the learning rate \cite{smith2017don}. In all experiments (and for all models), the corpus is split into 50\% training sentences and 50\% test sentences. The reported results always refer to the results on the test set.

\subsection{Analysis Model}

The analysis model is used to analyze what information is stored in the internal representation of the LSTM. In a Push-Down-Automaton model, this internal representation would conceptually correspond to the entire stack.

To analyze the internal representation $[h_i, c_i]$ of the LSTM after having read the input or part of it, we use a method already developed by \citeauthor{shi2016does} \shortcite{shi2016does} and \citeauthor{belinkov2017neural} \shortcite{belinkov2017neural}: After having trained the basic model, the weights of the encoder are fixed and the labels (previously $y$) are replaced by some feature $z_i$ of the input $x_1,\ldots, x_i$.

This feature $z_i$ can either be a scalar or a vector. If $z_i$ is a scalar, a dense layer (scalar analysis decoder) is trained to predict $z_i$. On the other hand, if $z_i$ is a vector (sequence analysis decoder), another LSTM is trained to predict $z_{i,1},\ldots,z_{i,j}$.\\

Analyzing the performance of the analysis network shows us how accurately a feature $z_i$ is preserved in $[h_i, c_i]$. One can assume that the LSTM uses its limited memory ``efficiently'' and therefore discards irrelevant information. Hence, the performance of the analysis decoder shows whether $z_i$ is contained in the information that is relevant for the original classification task.

To begin, two of the experiments which were conducted are presented in the following section to test the trained model performance. For the first experiment $z_i$ is the \emph{depth} (nesting level) after $i$ characters.

\textbf{Example:} For the sequence \verb|{[{}[[]|, $z$ is $(1,2,3,2,3,4,3)$.\\

For the second experiment we note that theoretically, at any time $t$, no information about a closed clause in $w_1,\ldots, w_t$ has to be stored, since it is irrelevant for any eventual future prediction of $w_{t+1}, w_{t+2},\ldots$. 
When reading from left to right, as soon as a closing bracket is processed, the corresponding clause becomes irrelevant. Therefore, the relevant information is simply the list of bracket types of unclosed clauses. In this experiment we investigate how well the previous characters are preserved in the intermediate representation and evaluate if this correlates with the recovered characters being relevant or not.

\textbf{Example:} after having processed \verb|{[{}][|, only the first and the last characters are relevant, since they are the only ones that could matter for a future classification task. On the other hand, the sub-string \verb|[{}]| is irrelevant. In this example we would evaluate whether the first and last character are better preserved in the intermediate state than the irrelevant sub-string.

To set up the experiment, we set $z_{i,k}$ to be equal to $x_{i-k+1}$, corresponding to predicting the previous characters of a given intermediate state.

\subsection{Varying hidden units}
\label{subsec:varying_hidden_units}

The basic model is evaluated with $2,4,6,...,50$ hidden units. The error rate with 50 hidden units is 0.38\% and an error rate of 1\% is reached around 20 hidden units.  Thus, the error seems to converge with increasing hidden units to a fairly small value. As a result, in all further experiments, the maximum number of hidden units the models are tested against was set to 50.

\setlength\figureheight{4cm}
\setlength\figurewidth{\linewidth}
\begin{figure}[ht]
    \input{figures/varying_units_results.tex}%
    \caption{Overall error rate of the basic model with respect to the number of hidden units of the LSTM.}%
    \label{fig:varying_units_results}%
\end{figure}
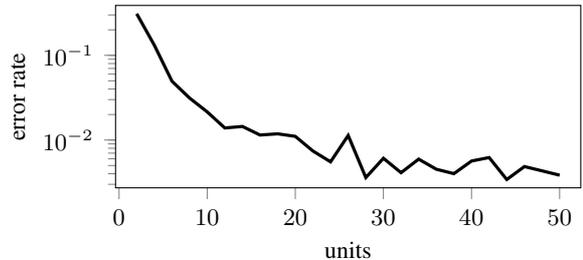

\subsection{Memory demand}
\label{subsec:memory_demand}

In this section we evaluate how ``difficult'' sentences can be with respect to the memory demand of the model, while still reaching an error tolerance of 5\%. We have to work with tolerances, because 100\% accuracy is not reached. Since it can be challenging measuring how difficult a sentence is to predict, we use the \emph{distance} and the \emph{embedded depth} of a sentence as defined above as metrics.

\setlength\figureheight{5cm}
\setlength\figurewidth{0.48\textwidth}
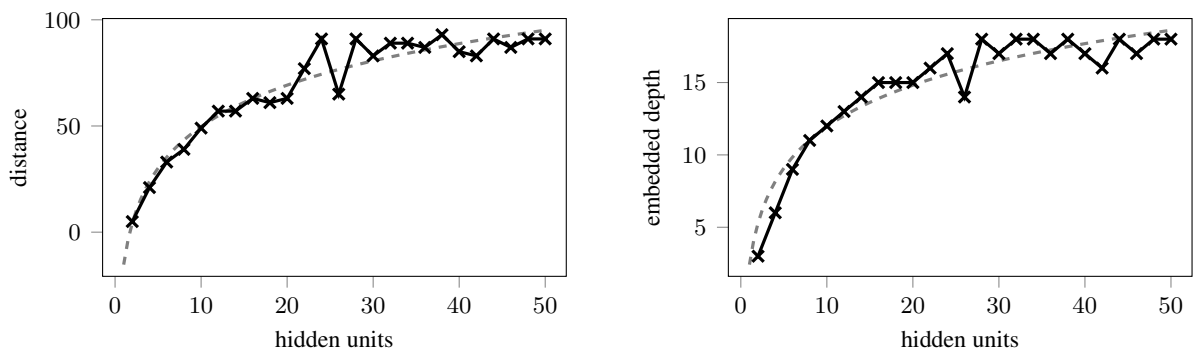
\begin{figure*}[ht]
    \input{figures/memory_demand/distance.tex}\qquad
    \input{figures/memory_demand/depth_increase.tex}
    \caption{\emph{Distances} (left figure) and \emph{embedded depth} (right figure) that can be predicted with a given number of hidden units and 5\% error tolerance. The dashed line is a logarithmic approximation.}%
    \label{fig:memory_demand}%
\end{figure*}

The resulting values (figure \ref{fig:memory_demand}) demonstrate that memory demand grows exponentially with respect to the \emph{distance} of sentences that can be predicted. The same behaviour can be observed with respect to the \emph{embedded depth}. 

\subsection{Generalization}
\label{subsec:generalization}

To evaluate the model's generalization performance, training was done only on a systematically chosen subset of sentences (in-sample). To avoid adding additional nuisance variables in this selection, the training sentences are selected according to one of the following rules:

\begin{samepage}
\begin{description}
    \item[regular interpolation:] the sentence has \emph{distance} $2,6,10,14,\ldots$ / odd \emph{embedded depth}.
    \item[random interpolation:] the \emph{distance} / \emph{embedded depth} of the relevant clause belongs to a set $D$. $D$ is a random subset consisting of half of all \emph{distances} / \emph{embedded depths} present in the corpus.
    \item[extrapolation:] the \emph{distance} / \emph{embedded depth} of the relevant clause is smaller than a certain threshold (11 for \emph{distance} and 13 for \emph{embedded depth}).
\end{description}
\end{samepage}

\begin{figure}
    \centering
    \input{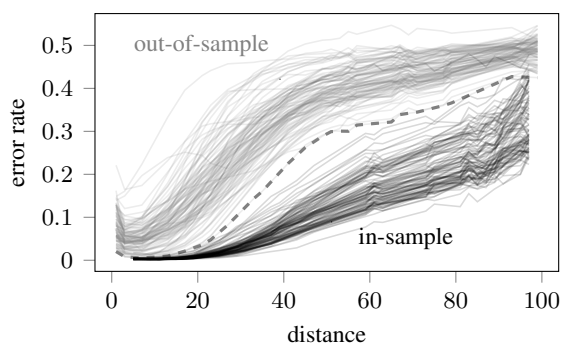}
    \caption{Generalization behaviour of a model with 10 hidden units with respect to the \emph{distance} for 100 runs. Half of the corpus is systematically selected for training (in-sample) -- for testing also the left out \emph{distances} are considered (out-of-sample). The bold dashed line is the minimal out-of-sample error out of all 100 runs.}
    \label{fig:distance_random_multiple}
\end{figure}

Running the experiment 100 times -- each one with a different random weight initialization -- has shown (figure \ref{fig:distance_random_multiple}) that the results are consistent with respect to the weight initialization. The best out-of-sample accuracy is still worse than almost all in-sample accuracies.

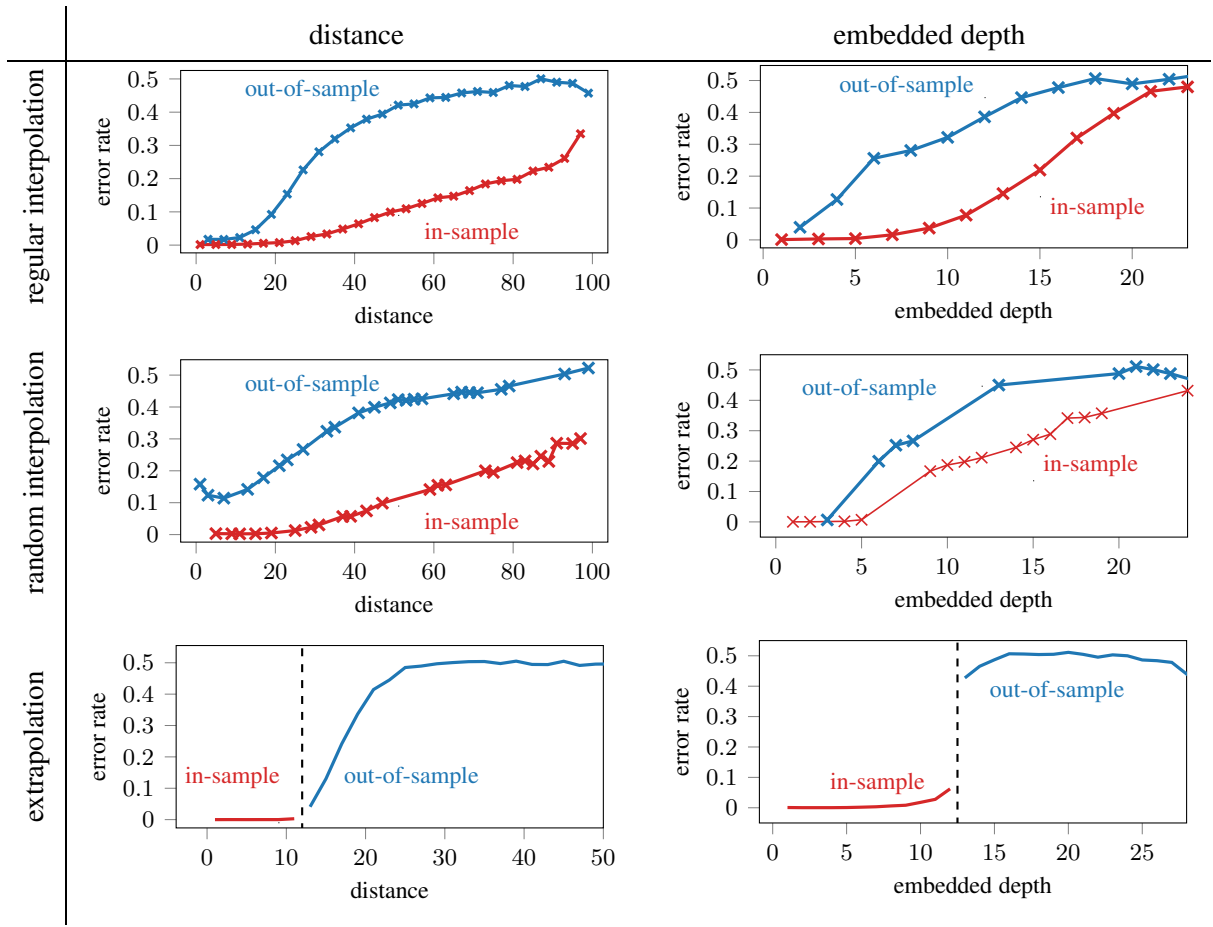
\begin{figure*}[ht]
    \bgroup
    \def\arraystretch{1.5}%
    \begin{tabular}{c|c c}
        & distance & embedded depth\\
        \hline
        \raisebox{0.15in}{\rotatebox{90}{regular interpolation}} &
        \input{figures/generalization/distance_even_odd.tex} & \input{figures/generalization/depth_increase_even_odd.tex}\\
        \raisebox{0.15in}{\rotatebox{90}{random interpolation}} &
        \input{figures/generalization/distance_random.tex} & \input{figures/generalization/depth_increase_random.tex}\\
        \raisebox{0.45in}{\rotatebox{90}{extrapolation}} & \input{figures/generalization/distance_extrapolation.tex} &
        \input{figures/generalization/depth_increase_extrapolate.tex}
    \end{tabular}
    \egroup
    \caption{Test for generalization: The error rate of the model with 10 hidden units if only half of the corpus is systematically selected for training (in-sample), while during testing also the left out \emph{distances} / \emph{embedded depths} were considered (out-of-sample).}%
    \label{fig:generalization_results}%
\end{figure*}

The results (figure \ref{fig:generalization_results}) demonstrate a large discrepancy between the performance on in-sample (training) and the out-of-sample (testing) accuracy. The experiment was evaluated for different numbers of hidden units. On the one hand, with a large number of hidden units, the generalization error is similarly large (the out-of-sample error rate for interpolation was already between 8.1 and 14.3 times larger than the in-sample error). On the other hand, models with a small number of hidden units did not even converge. The reason for no convergence can be reasonably be explained by the sparse data set, that might lead to more local minima. The maximum generality -- especially for smaller \emph{distances} -- is observed at around 10 hidden units.

Generalization was evaluated with respect to \emph{distance} and with respect to the \emph{embedded depth}.

For regular interpolation the out-of-sample error for 10 hidden units was on average 5.4 (\emph{distance}) and 5.9 (\emph{embedded depth}) times higher than the in-sample error. Figure \ref{fig:generalization_results} shows also that for random interpolation and extrapolation, the model generalizes much worse or not at all.

\setlength\figureheight{4cm}
\setlength\figurewidth{0.45\textwidth}

\subsection{Intermediate State Analysis}
\label{subsec:intermediate_state_analysis}

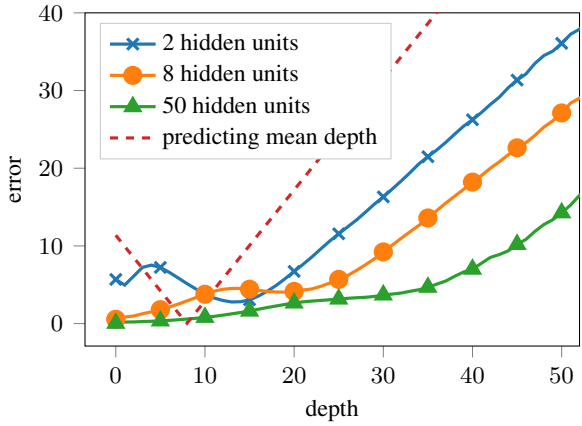
\begin{figure}[ht]
    \input{figures/analysis_depth_results.tex}%
    \caption{Error rates of predicting the \emph{depth} given an intermediate state of the basic model. It is evaluated for $2,8,50$ hidden units. The dashed line indicates the baseline always predicting the mean \emph{depth}. The error indicates how far the prediction is off from the true \emph{depth}.}%
    \label{fig:analysis_depth_results}%
\end{figure}

For the first experiment analyzing intermediate states, recovering the \emph{depth} as defined in section \ref{subsec:basic_model} from intermediate states shows that the \emph{depth} is only marginally conserved in the intermediate state (figure \ref{fig:analysis_depth_results}). For a small number of units, the model is only able to distinguish whether a \emph{depth} is either close to $0$ or close to the mean \emph{depth} (see figure \ref{fig:analysis_depth_results} with two hidden units). When increasing the number of hidden units, the distribution of predictions gets closer to the real distribution of \emph{depths}.

While for two hidden units the prediction of the \emph{depth} is on average off by $7.04$, it decreases until it reaches a value of $1.34$ for 50 hidden units.

\setlength\figureheight{6cm}
\setlength\figurewidth{1.05\linewidth}

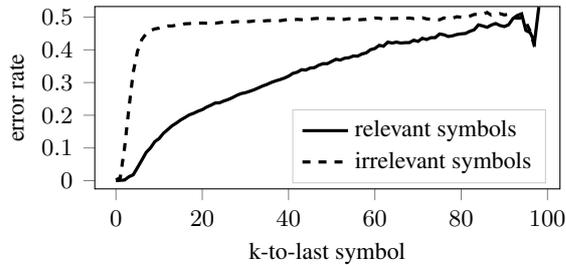
\begin{figure}[ht]
    \input{figures/analysis_previous_results.tex}%
    \caption{Error rate of predicting the previous characters given an intermediate state of the basic model with 20 hidden units. Characters are grouped as being either relevant or irrelevant for the basic classification task.}%
    \label{fig:analysis_previous_results}%
\end{figure}

Figure \ref{fig:analysis_previous_results} shows how accurately a past character can be recovered from an intermediate state. There is a large discrepancy between the accuracy of relevant and irrelevant characters: If the 4th-to-last character is an irrelevant one, the model is only able to recover the type of bracket with a 33\% error rate; whereas if it is a relevant character, it reaches an error below 1.8\%. As the number of past characters $k$ approaches 10, the irrelevant information cannot be recovered anymore. Note that an error rate of $0.5$ amounts to a random guess, since we evaluate only if it can predict the type of bracket (square or curly) correctly, and not whether it was an opening or a closing one.

\setlength\figureheight{4cm}
\setlength\figurewidth{\linewidth}

\section{Discussion}

We now consider the results of the various experiments, some of which might be considered as controversial on first sight. On the one hand, we see that the LSTM exhibits an exponential memory demand as sentences grow longer, while theoretically, a sub-linear memory ought to be sufficient \cite{magniez2014recognizing}. On the other hand, we see that the model has successfully sorted out irrelevant information: the intermediate state analysis shows that irrelevant characters are very quickly forgotten. So, the exponential memory space is not needed for storing irrelevant information for the original classification task.

The strength of structural rules is that they generalize well. In human language this enables humans to create new sentences which have never been heard before. But also for the Dyck language being used, the 4 rules defining the language are enough to generate sentences of arbitrary \emph{distance} and \emph{(embedded) depth}. The only constraint is the memory to store intermediate results while streaming the input. Assuming the model had in fact learned the underlying grammatical rules correctly, an upper bound for the memory required is 50 bits. The model we are using has up to 50 hidden units which corresponds to 11,200 trainable parameters. \citeauthor{collins2016capacity} \shortcite{collins2016capacity} showed that LSTMs can store up to 5 bits of information per parameter and one real number per hidden unit. So we can assume that memory to store the values to process the corpus-defining rules sequentially is not an issue. To partially answer the question of whether LSTM can learn rules we follow a proof by contradiction: if the LSTM learns rules and if these rules are the correct ones, the model would generalize. What we observe is that the LSTM generalizes poorly. Therefore we conclude that the model is not able to learn the right rules.

Combining the generalization results and the intermediate state analysis reveals that the model determines each character's relevance -- but it has learned this without resorting to hierarchical rules. As LSTMs are known to have the ability to capture statistical contingencies, it suggests instead that rather than the ``perfect'' rule-based solution, what the LSTM has in fact acquired is a sequential statistical approximation to this solution.

The large effect of initialization to a good local minimum suggests that the underlying function may well have many local minima as on reviewers noted. Indeed, \citeauthor{collins2016capacity} \shortcite{collins2016capacity} has already concluded that the memory in LSTMs is mainly used for training effectiveness rather than to increase the storage capacity. Therefore, the large memory demand in our experiments suggests that the LSTM memory is needed to avoid such local minima.

\section{Conclusion}

At heart, neural networks are statistical models, performing well at capturing and combining correlations of the output variable values and the corresponding component values in the training input. In particular, LSTMs are constructed such that they capture sequential information. Hence, due to the design of their architecture, LSTMs perform very well on statistically-oriented, sequential tasks. 

As a result, in experiments like this one that examine whether LSTMs can acquire hierarchical knowledge, one has to pay close attention to nuisance variables like sequential statistical correlations that might be hard to detect and confounded with true hierarchical information.

The bottom line that emerges from this experiment is that the range of rules that an LSTM can learn is very restricted: even a context-free grammar with four simple rules apparently cannot be appropriately learned by an LSTM. 

According to most linguistic accounts, natural language syntax relies heavily on hierarchical rules. It enables humans to compose new sentences with relatively little memory capacity and training data. Furthermore, there are sentences that have the same linear representation but differ in structure -- syntactically ambiguous sentences.  From this perspective, it seems not only more efficient to directly infer structures and rules, but also useful to use rules to understand sentences correctly. The bracket completion task presented here can be understood by a human after only a few training sentences, though online processing of the rules themselves may be difficult. This result invites the conclusion that it will be very challenging for LSTMs to understand natural language as humans do. While LSTMs remain good engineering tools to approximate certain language features based on statistical correlations, the exploration of fundamentally new models and architectures seems a valuable direction to explore on the way to developing methods for understanding human language in the way that people do.

\section*{Acknowledgments}
We want to thank Beracah Yankama for his help, valuable discussions and the machines to run the experiments on. We thank Prof. Thomas Hofmann for the fast and easy administrative process at ETH Zurich and also for granting access to high-computing clusters. Additionally we are very grateful for the financial support provided by the Zeno Karl Schindler Foundation.

\bibliography{main}
\bibliographystyle{acl_natbib_nourl}

\end{document}

%% file: figures/corpus_distance_frequencies.tex
% This file was created by matplotlib2tikz v0.6.17.
\begin{tikzpicture}

\definecolor{color0}{rgb}{0.12156862745098,0.466666666666667,0.705882352941177}

\begin{axis}[
xlabel={distance},
ylabel={frequency},
xmin=-3.9, xmax=103.9,
ymin=0, ymax=491817.9,
width=\figurewidth,
height=\figureheight,
tick align=outside,
tick pos=left,
x grid style={lightgray!92.02614379084967!black},
y grid style={lightgray!92.02614379084967!black}
]
\draw[fill=color0,draw opacity=0] (axis cs:1,0) rectangle (axis cs:10.8,468398);
\draw[fill=color0,draw opacity=0] (axis cs:10.8,0) rectangle (axis cs:20.6,143351);
\draw[fill=color0,draw opacity=0] (axis cs:20.6,0) rectangle (axis cs:30.4,98107);
\draw[fill=color0,draw opacity=0] (axis cs:30.4,0) rectangle (axis cs:40.2,83418);
\draw[fill=color0,draw opacity=0] (axis cs:40.2,0) rectangle (axis cs:50,66428);
\draw[fill=color0,draw opacity=0] (axis cs:50,0) rectangle (axis cs:59.8,49931);
\draw[fill=color0,draw opacity=0] (axis cs:59.8,0) rectangle (axis cs:69.6,36258);
\draw[fill=color0,draw opacity=0] (axis cs:69.6,0) rectangle (axis cs:79.4,26729);
\draw[fill=color0,draw opacity=0] (axis cs:79.4,0) rectangle (axis cs:89.2,18836);
\draw[fill=color0,draw opacity=0] (axis cs:89.2,0) rectangle (axis cs:99,8544);
\end{axis}

\end{tikzpicture}

%% file: figures/corpus_depth_frequencies.tex
% This file was created by matplotlib2tikz v0.6.17.
\begin{tikzpicture}

\definecolor{color0}{rgb}{0.12156862745098,0.466666666666667,0.705882352941177}

\begin{axis}[
xlabel={depth},
ylabel={frequency},
xmin=-3.5, xmax=73.5,
ymin=0, ymax=408977.1,
width=\figurewidth,
height=\figureheight,
tick align=outside,
tick pos=left,
x grid style={lightcolor0!92.02614379084967!black},
y grid style={lightcolor0!92.02614379084967!black}
]
\draw[fill=color0,draw opacity=0] (axis cs:0,0) rectangle (axis cs:7,389502);
\draw[fill=color0,draw opacity=0] (axis cs:7,0) rectangle (axis cs:14,216240);
\draw[fill=color0,draw opacity=0] (axis cs:14,0) rectangle (axis cs:21,152435);
\draw[fill=color0,draw opacity=0] (axis cs:21,0) rectangle (axis cs:28,108533);
\draw[fill=color0,draw opacity=0] (axis cs:28,0) rectangle (axis cs:35,72311);
\draw[fill=color0,draw opacity=0] (axis cs:35,0) rectangle (axis cs:42,40144);
\draw[fill=color0,draw opacity=0] (axis cs:42,0) rectangle (axis cs:49,16378);
\draw[fill=color0,draw opacity=0] (axis cs:49,0) rectangle (axis cs:56,3966);
\draw[fill=color0,draw opacity=0] (axis cs:56,0) rectangle (axis cs:63,471);
\draw[fill=color0,draw opacity=0] (axis cs:63,0) rectangle (axis cs:70,20);
\end{axis}

\end{tikzpicture}

%% file: figures/lstm_cell.tex
\begin{tikzpicture}[auto, thick, node distance=1cm, align=center, inner sep=2]

\draw
    (2.2,0) node [name=input] {$x_t$};
\coordinate[above=of input] (p1);
\draw
    node [right=1.2cm of p1, circle, draw, name=input activation, inner sep=1, scale=0.8] {$tanh$}
    node [above of=input activation, name=input multiplication] {$\bigodot$}
    node [above of=input multiplication, name=addition] {$\bigoplus$}
    node [right of=addition, circle, draw, name=memory, node distance=0.9cm] {$c_t$}
    node [above of=memory, circle, draw, name=output activation, inner sep=1, scale=0.8] {$tanh$}
    node [above of=output activation, name=output multiplication] {$\bigodot$}
    node [above of=output multiplication, name=output] {$h_t$}
    
    node[left of=input multiplication, circle, draw, name=input gate, node distance=0.9cm] {$i_t$}
    node[left of=addition, name=memory multiplication, node distance=0.8cm] {$\bigodot$}
    node[above of=memory multiplication, circle, draw, name=forget gate, node distance=0.9cm] {$f_t$}
    node[left of=output multiplication, circle, draw, name=output gate, node distance=0.9cm] {$o_t$}

;

\coordinate (prev output x) at (0.8,0);
\draw
    (0,0) node [name=prev input] {$x_{t-1}$}
    node [above of=prev input, node distance=1.5cm, name=input ellipsis] {$\vdots$}
    (prev output x |- output) node [name=prev output] {$h_{t-1}$}
    (prev output.south |- memory multiplication) node [circle, draw, name=prev memory, inner sep=0] {$c_{t-1}$}
    node [left of=prev memory, node distance=0.9cm, name=memory ellipsis in, scale=0.8] {$\ldots$}
    node [above of=prev memory, node distance=1.2cm, name=memory ellipsis out] {$\vdots$}

    node [below of=prev output, node distance=1.2cm, name=output ellipsis] {$\vdots$}
;

\draw
    (6.2,0) node [name=next input] {$x_{t+1}$}
    node [above of=next input, node distance=1.5cm, name=next input ellipsis] {$\vdots$}
    (next input |- memory) node [name=next memory ellipsis, scale=0.8] {$\ldots$}
    (next input |- output) node [yshift=-1cm, node distance=3.8cm, name=next output ellipsis] {$\vdots$}
;

\coordinate (p2) at (p1 |- input gate);
\coordinate (p3) at (p2 |- forget gate);
\coordinate (p4) at (p3 |- output gate);
\coordinate (p5) at ([xshift=-0.2cm]p4 |- prev output);

\draw[->] (prev input) -- (input ellipsis);
\draw[->, shorten >=1pt] (memory ellipsis in) -- (prev memory);
\draw[->, shorten <=2pt] (prev memory) -- (memory ellipsis out);
\draw[->] (output ellipsis) -- (prev output);

\draw[->, shorten <=2pt, shorten >=-2pt] (prev memory) -- (memory multiplication);

\draw[->, shorten >=1pt] (prev output) -- (p5) -- ([yshift=0.1cm, xshift=-0.2cm]p4) -- ([yshift=0.1cm]output gate.west);
\draw[->, shorten >=1pt] ([yshift=0.1cm, xshift=-0.2cm]p4) -- ([yshift=0.1cm, xshift=-0.2cm]p3) -- ([yshift=0.1cm]forget gate.west);
\draw[->, shorten >=1pt] ([yshift=0.1cm, xshift=-0.2cm]p3) -- ([yshift=0.1cm, xshift=-0.2cm]p2) -- ([yshift=0.1cm]input gate.west);
\draw[->, shorten >=1pt] ([yshift=0.1cm, xshift=-0.2cm]p2) -- ([yshift=0.1cm, xshift=-0.2cm]p1) -- ([yshift=0.1cm]input activation.west);

\draw[->, shorten >=1pt] (input) -- ([yshift=-0.1cm]p1) -- ([yshift=-0.1cm]input activation.west);
\draw[->, shorten >=1pt] ([yshift=-0.1cm]p1) -- ([yshift=-0.1cm]p2) -- ([yshift=-0.1cm]input gate.west);
\draw[->, shorten >=1pt] ([yshift=-0.1cm]p2) -- ([yshift=-0.1cm]p3) -- ([yshift=-0.1cm]forget gate.west);
\draw[->, shorten >=1pt] ([yshift=-0.1cm]p3) -- ([yshift=-0.1cm]p4) -- ([yshift=-0.1cm]output gate.west);
\draw[->, shorten >= -2pt] (input activation) -- (input multiplication);
\draw[->, shorten <=-2pt, shorten >=-2pt] (input multiplication) -- (addition);
\draw[->, shorten >=1pt, shorten <= -2pt] (addition) -- (memory);
\draw[->, shorten >=1pt, shorten <=1pt] (memory) -- (output activation);
\draw[->, shorten <=1pt] (output activation) -- (output multiplication);
\draw[->, shorten <=1pt] (output multiplication) -- (output);

\draw[->, shorten <=1pt, shorten >=-3pt] (input gate) -- (input multiplication);
\draw[->, shorten <=-2pt, shorten >= -2pt] (memory multiplication) -- (addition);
\draw[->, shorten <=1pt, shorten >=-3pt] (forget gate) -- (memory multiplication);
\draw[->, shorten <=1pt, shorten >=-3pt] (output gate) -- (output multiplication);

%% Boxes

\draw[->] (output) -- (next output ellipsis |- output) -- ([yshift=-0.2cm]next output ellipsis.north);
\draw[->, shorten <= 2pt] (memory) -- (next memory ellipsis);
\draw[->] (next input) -- (next input ellipsis);

\draw[gray] (1.7,0.7) rectangle (5.3,5.8);

\draw[dotted, gray] (-0.4,0.7) rectangle (1.4,5.8);

\draw[dotted, gray] (5.6,0.7) rectangle (6.9,5.8);

\end{tikzpicture}

%% file: figures/lstm_architecture.tex
    \begin{tikzpicture}[node distance=1.7cm, every text node part/.style={align=center}]
    \node (input1) at (0,0) {$x_1$};
    \node[above=0.5cm of input1, draw] (embedding1) {Embedding};
    \node[above=0.5cm of embedding1, draw] (lstm1) {LSTM \\ cell};

    \node[right=4cm of input1] (input3) {$x_i$};
    \node[above=0.5cm of input3, draw] (embedding3) {Embedding};
    \node[above=0.5cm of embedding3, draw] (lstm3) {LSTM \\ cell};
    \node[above of=lstm3, draw] (dense3) {Dense};
    \node[above=0.5cm of dense3] (output3) {$y_i$};
    
    \node[right=1cm of dense3, draw] (dense4) {Dense};
    \node[above=0.5cm of dense4] (output4) {$z_i$};
    
    \node[right=3cm of lstm3, draw] (lstm5) {LSTM \\ cell};
    \node[above of=lstm5, draw] (dense5) {Dense};
    \node[above=0.5cm of dense5] (output5) {$z_{i,1}$};

    \node[right=3cm of lstm5, draw] (lstm6) {LSTM \\ cell};
    \node[above of=lstm6, draw] (dense6) {Dense};
    \node[above=0.5cm of dense6] (output6) {$z_{i,j}$};
    
    \draw[->, shorten >= 1pt] (input1) -- (embedding1);
    \draw[->, shorten >= 1pt, shorten <= 1pt] (embedding1) -- (lstm1);

    \node (lstm dots) at ($(lstm1)!0.5!(lstm3)$) {$\ldots$};
    
    \draw[->, shorten <= 1pt] (lstm1) -- node[above]{$h_1$} node[below]{$c_1$} ++ (lstm dots);
    \draw[->, shorten <= 1pt] (lstm dots) -- node[above]{$h_{i-1}$} node[below]{$c_{i-1}$} ++ (lstm3);

    \draw[->, shorten >= 1pt] (input3) -- (embedding3);
    \draw[->, shorten >= 1pt, shorten <= 1pt] (embedding3) -- (lstm3);
    \draw[->, shorten >= 1pt, shorten <= 1pt] (lstm3) -- node[right] {$h_i$} ++ (dense3);
    \draw[->, shorten <= 1pt] (dense3) -- (output3);
    
    \draw[->, shorten <= 1pt, shorten >= 1pt] (lstm3) -- node[right] {$h_i, c_i$} ++ (dense4);
    \draw[->, shorten <= 1pt] (dense4) -- (output4);
    
    \draw[->, shorten <= 1pt, shorten >= 1pt] (lstm3) -- node[above] {$h_i$} node[below] {$c_i$} ++ (lstm5);
    
    \node[blue] (lstm dots analysis) at ($(lstm5)!0.5!(lstm6)$) {$\ldots$};
    
    \draw[->, shorten <= 1pt] (lstm5) -- node[above]{$h'_1$} node[below]{$c'_1$} ++ (lstm dots analysis);
    \draw[->, shorten <= 1pt] (lstm dots analysis) -- node[above]{$h'_{j-1}$} node[below]{$c'_{j-1}$} ++ (lstm6);
    \draw[->, shorten <= 1pt, shorten >= 1pt] (lstm5) -- node[right] {$h'_1$} ++ (dense5);
    \draw[->, shorten <= 1pt, shorten >= 1pt] (lstm6) -- node[right] {$h'_j$} ++ (dense6);
    \draw[->, shorten <= 1pt] (dense5) -- (output5);
    \draw[->, shorten <= 1pt] (dense6) -- (output6);
    
    \draw (-1.2,-0.3) rectangle (5.9,3.1);
    \draw (-1,3.1) node[anchor=south west] {encoder};
    
    \draw[dashed] (3.8, 3.6) rectangle (5.5, 5.5);
    \draw (4.65,5.5) node[anchor=south] {basic \\ decoder};
    
    \draw[dashed] (6, 3.6) rectangle (7.7, 5.5);
    \draw (6.85, 5.5) node[anchor=south] {scalar analysis \\ decoder};
    
    \draw[dashed] (8.1, 1.6) rectangle (14.1, 5.5);
    \draw (11.1, 5.5) node[anchor=south] {sequence analysis decoder};
    
    \end{tikzpicture}

%% file: figures/varying_units_results.tex
% This file was created by matplotlib2tikz v0.6.17.
\begin{tikzpicture}

\definecolor{color0}{rgb}{0.12156862745098,0.466666666666667,0.705882352941177}

\begin{axis}[
xlabel={units},
ylabel={error rate},
xmin=-0.4, xmax=52.4,
ymin=0.00272669964390025, ymax=0.388158976867021,
ymode=log,
width=\figurewidth,
height=\figureheight,
tick align=outside,
tick pos=left,
x grid style={lightgray!92.02614379084967!black},
y grid style={lightgray!92.02614379084967!black}
]
\addplot [very thick, black, forget plot]
table {%
2 0.309834
4 0.131702
6 0.0495
8 0.0313
10 0.02149
12 0.0138779999999999
14 0.014466
16 0.011456
18 0.011836
20 0.01105
22 0.00742200000000004
24 0.00552600000000003
26 0.011324
28 0.00360400000000005
30 0.00609000000000004
32 0.004112
34 0.00595000000000001
36 0.00451999999999997
38 0.00400800000000001
40 0.00564799999999999
42 0.006212
44 0.00341599999999997
46 0.00486200000000003
48 0.00432600000000005
50 0.00384399999999996
};
\end{axis}

\end{tikzpicture}

%% file: figures/memory_demand/distance.tex
% This file was created by matplotlib2tikz v0.6.17.
\begin{tikzpicture}

\definecolor{color0}{rgb}{0.12156862745098,0.466666666666667,0.705882352941177}

\begin{axis}[
xlabel={hidden units},
ylabel={distance},
xmin=-1.45, xmax=52.45,
ymin=-20.802587179018, ymax=100.597830773957,
width=\figurewidth,
height=\figureheight,
tick align=outside,
tick pos=left,
x grid style={lightgray!92.02614379084967!black},
y grid style={lightgray!92.02614379084967!black}
]
\addplot [very thick, gray, dashed, forget plot]
table {%
1 -15.2843863629737
1.4949494949495 -3.94075808103901
1.98989898989899 4.1274883379046
2.48484848484848 10.3940457953906
2.97979797979798 15.5184894738201
3.47474747474747 19.8536455124497
3.96969696969697 23.6105125561782
4.46464646464647 26.9253708075226
4.95959595959596 29.8913650243448
5.45454545454546 32.5749837904557
5.94949494949495 35.0253508303714
6.44444444444444 37.2797906080873
6.93939393939394 39.3673286210571
7.43434343434343 41.3109828660492
7.92929292929293 43.129315234753
8.42424242424243 44.8375123408373
8.91919191919192 46.4481573049852
9.41414141414142 47.9717928239192
9.90909090909091 49.417339794359
10.4040404040404 50.7924137959027
10.8989898989899 52.1035679524578
11.3939393939394 53.3564818131212
11.8888888888889 54.5561100396219
12.3838383838384 55.7068007463889
12.8787878787879 56.812390635396
13.3737373737374 57.8762821806949
13.8686868686869 58.9015067795721
14.3636363636364 59.8907768249906
14.8585858585859 60.8465289527432
15.3535353535354 61.7709601994918
15.8484848484848 62.6660584220042
16.3434343434343 63.533628037042
16.8383838383838 64.3753119199611
17.3333333333333 65.1926101300394
17.8282828282828 65.9868959988224
18.3232323232323 66.7594300149295
18.8181818181818 67.5113718578532
19.3131313131313 68.2437908691913
19.8080808080808 68.957675198637
20.3030303030303 69.6539398210373
20.7979797979798 70.333433587715
21.2929292929293 70.9969454483795
21.7878787878788 71.6452099580086
22.2828282828283 72.2789121650987
22.7777777777778 72.8986919628477
23.2727272727273 73.5051479725595
23.7676767676768 74.0988410183493
24.2626262626263 74.6802972437022
24.7575757575758 75.2500109132899
25.2525252525253 75.8084469374343
25.7474747474747 76.3560431515272
26.2424242424242 76.8932123784037
26.7373737373737 77.4203442980101
27.2323232323232 77.9378071455759
27.7272727272727 78.4459492568276
28.2222222222222 78.9451004764848
28.7171717171717 79.4355734443031
29.2121212121212 79.9176647712193
29.7070707070707 80.3916561166812
30.2020202020202 80.8578151769584
30.6969696969697 81.3163965931191
31.1919191919192 81.7676427863818
31.6868686868687 82.2117847277047
32.1818181818182 82.6490426477279
32.6767676767677 83.0796266925323
33.1717171717172 83.5037375301023
33.6666666666667 83.9215669118738
34.1616161616162 84.3332981933016
34.6565656565657 84.7391068169815
35.1515151515152 85.1391607615167
35.6464646464647 85.5336209589987
36.1414141414141 85.9226416837032
36.6363636363636 86.3063709143483
37.1313131313131 86.6849506720443
37.6262626262626 87.058517335867
38.1212121212121 87.4272019378091
38.6161616161616 87.7911304387056
39.1111111111111 88.1504239865873
39.6060606060606 88.5051991587889
40.1010101010101 88.8555681890217
40.5959595959596 89.2016391805178
41.0909090909091 89.5435163062591
41.5858585858586 89.8812999972179
42.0808080808081 90.2150871194614
42.5757575757576 90.5449711409009
43.0707070707071 90.8710422884036
43.5656565656566 91.1933876959289
44.0606060606061 91.5120915442953
44.5555555555556 91.8272351931423
45.0505050505051 92.1388973056008
45.5454545454545 92.4471539661538
46.040404040404 92.7520787921258
46.5353535353535 93.0537430392123
47.030303030303 93.3522157014257
47.5252525252525 93.6475636058101
48.020202020202 93.9398515022483
48.5151515151515 94.2291421486653
49.010101010101 94.5154963919065
49.5050505050505 94.7989732445524
50 95.0796299579128
};
\addplot [very thick, black, mark=x, mark size=3, mark options={solid}, forget plot]
table {%
2 5
4 21
6 33
8 39
10 49
12 57
14 57
16 63
18 61
20 63
22 77
24 91
26 65
28 91
30 83
32 89
34 89
36 87
38 93
40 85
42 83
44 91
46 87
48 91
50 91
};
\end{axis}

\end{tikzpicture}

%% file: figures/memory_demand/depth_increase.tex
% This file was created by matplotlib2tikz v0.6.17.
\begin{tikzpicture}

\definecolor{color0}{rgb}{0.12156862745098,0.466666666666667,0.705882352941177}

\begin{axis}[
xlabel={hidden units},
ylabel={embedded depth},
xmin=-1.45, xmax=52.45,
ymin=1.6079327425331, ymax=19.4243492868153,
width=\figurewidth,
height=\figureheight,
tick align=outside,
tick pos=left,
x grid style={lightgray!92.02614379084967!black},
y grid style={lightgray!92.02614379084967!black}
]
\addplot [very thick, gray, dashed, forget plot]
table {%
1 2.41776985818229
1.4949494949495 4.08253189114937
1.98989898989899 5.2666072131758
2.48484848484848 6.1862712460006
2.97979797979798 6.93832156569536
3.47474747474747 7.57453803900771
3.96969696969697 8.12588629036611
4.46464646464647 8.61236645564922
4.95959595959596 9.04764822259756
5.45454545454546 9.4414892866506
5.94949494949495 9.80109892066709
6.44444444444444 10.1319547680444
6.93939393939394 10.4383165369593
7.43434343434343 10.7235622900692
7.92929292929293 10.990416123384
8.42424242424243 11.2411067856973
8.91919191919192 11.4774809437939
9.41414141414142 11.7010858170284
9.90909090909091 11.9132306135122
10.4040404040404 12.115032976336
10.8989898989899 12.3074546252961
11.3939393939394 12.4913290765713
11.8888888888889 12.6673834637143
12.3838383838384 12.8362559049388
12.8787878787879 12.9985094648675
13.3737373737374 13.1546434819375
13.8686868686869 13.3051028363042
14.3636363636364 13.4502855918593
14.8585858585859 13.590549343074
15.3535353535354 13.7262165214606
15.8484848484848 13.8575788598217
16.3434343434343 13.9849011697718
16.8383838383838 14.1084245555192
17.3333333333333 14.2283691619479
17.8282828282828 14.3449365357023
18.3232323232323 14.4583116628862
18.8181818181818 14.568664735112
19.3131313131313 14.6761526862326
19.8080808080808 14.7809205345836
20.3030303030303 14.8831025595461
20.7979797979798 14.9828233363811
21.2929292929293 15.0801986493413
21.7878787878788 15.1753362998468
22.2828282828283 15.2683368238724
22.7777777777778 15.3592941305161
23.2727272727273 15.4482960719179
23.7676767676768 15.5354249531985
24.2626262626263 15.6207579898373
24.7575757575758 15.7043677188605
25.2525252525253 15.7863223693243
25.7474747474747 15.866686196837
26.2424242424242 15.945519786228
26.7373737373737 16.022880325935
27.2323232323232 16.0988218572236
27.7272727272727 16.173395500959
28.2222222222222 16.2466496643132
28.7171717171717 16.3186302295016
29.2121212121212 16.3893807263907
29.7070707070707 16.4589424906048
30.2020202020202 16.5273548085674
30.6969696969697 16.594655050754
31.1919191919192 16.6608787942859
31.6868686868687 16.7260599358736
32.1818181818182 16.7902307960055
32.6767676767677 16.853422215186
33.1717171717172 16.9156636429381
33.6666666666667 16.9769832202147
34.1616161616162 17.0374078557961
34.6565656565657 17.0969632971912
35.1515151515152 17.1556741965127
35.6464646464647 17.2135641717449
36.1414141414141 17.2706558637881
36.6363636363636 17.3269709896227
37.1313131313131 17.3825303919055
37.6262626262626 17.4373540852827
38.1212121212121 17.4914612996764
38.6161616161616 17.5448705207782
39.1111111111111 17.5975995279651
39.6060606060606 17.6496654298302
40.1010101010101 17.7010846975073
40.5959595959596 17.751873195952
41.0909090909091 17.802046213326
41.5858585858586 17.8516184886235
42.0808080808081 17.9006042376619
42.5757575757576 17.9490171775543
43.0707070707071 17.9968705497662
43.5656565656566 18.0441771418558
44.0606060606061 18.0909493079859
44.5555555555556 18.1371989882902
45.0505050505051 18.1829377271699
45.5454545454545 18.22817669059
46.040404040404 18.2729266824419
46.5353535353535 18.3171981600301
47.030303030303 18.3610012487406
47.5252525252525 18.4043457559401
48.020202020202 18.4472411841569
48.5151515151515 18.4896967435845
49.010101010101 18.531721363951
49.5050505050505 18.5733237057927
50 18.6145121711661
};
\addplot [very thick, black, mark=x, mark size=3, mark options={solid}, forget plot]
table {%
2 3
4 6
6 9
8 11
10 12
12 13
14 14
16 15
18 15
20 15
22 16
24 17
26 14
28 18
30 17
32 18
34 18
36 17
38 18
40 17
42 16
44 18
46 17
48 18
50 18
};
\end{axis}

\end{tikzpicture}

%% file: figures/generalization/distance_even_odd.tex
% This file was created by matplotlib2tikz v0.6.17.
\begin{tikzpicture}

\definecolor{color0}{rgb}{0.12156862745098,0.466666666666667,0.705882352941177}
\definecolor{color1}{rgb}{0.83921568627451,0.152941176470588,0.156862745098039}

\begin{axis}[
xlabel={distance},
ylabel={error rate},
xmin=-3.9, xmax=103.9,
ytick={0,0.1,0.2,0.3,0.4,0.5},
ymin=-0.0233876745892656, ymax=0.525384728787867,
width=\figurewidth,
height=\figureheight,
tick align=outside,
tick pos=left,
x grid style={lightgray!92.02614379084967!black},
y grid style={lightgray!92.02614379084967!black}
]
\addplot [very thick, color0, forget plot, mark=x]
table {%
3 0.0174397202282348
7 0.0170740970927422
11 0.0227272727272727
15 0.0463058545937177
19 0.0924096814379783
23 0.153614155021615
27 0.226229335579657
31 0.280755314337404
35 0.319431797486797
39 0.352729178952886
43 0.378834132868149
47 0.394236909323116
51 0.421184919210054
55 0.424831336219917
59 0.442985487214927
63 0.444415104304199
67 0.457767722473605
71 0.462211748863828
75 0.459207014868471
79 0.480298189563365
83 0.477173385138417
87 0.500440528634361
91 0.490122996645546
95 0.486781609195402
99 0.457337883959044
};
\addplot [very thick, color1, forget plot, mark=x]
table {%
1 0.00155652556424046
5 0.00172546999742462
9 0.00189227248723811
13 0.00303490136570561
17 0.00552575979197134
21 0.0075700934579439
25 0.0131551472123617
29 0.0256667782613548
33 0.033531260915124
37 0.0483158824271477
41 0.0640526386464348
45 0.0827067669172933
49 0.0993746831164442
53 0.109369097091046
57 0.125134148959004
61 0.142182450907283
65 0.14721919302072
69 0.163995067817509
73 0.18397412863816
77 0.193649549196394
81 0.197988111568358
85 0.222916666666667
89 0.234281932495036
93 0.261019878997407
97 0.334943639291465
};
\draw[] (axis cs:49,0.421184919210054) -- (axis cs:49,0.421184919210054);
\node at (axis cs:49,0.401184919210054)[
  scale=1,
  anchor=south east,
  text=color0,
  rotate=0.0
]{ out-of-sample};
\draw[] (axis cs:51,0.0893746831164442) -- (axis cs:51,0.0893746831164442);
\node at (axis cs:55,0.0993746831164442)[
  scale=1,
  anchor=north west,
  text=color1,
  rotate=0.0
]{ in-sample};
\end{axis}

\end{tikzpicture}

%% file: figures/generalization/depth_increase_even_odd.tex
% This file was created by matplotlib2tikz v0.6.17.
\begin{tikzpicture}

\definecolor{color1}{rgb}{0.83921568627451,0.152941176470588,0.156862745098039}
\definecolor{color0}{rgb}{0.12156862745098,0.466666666666667,0.705882352941177}

\begin{axis}[
xlabel={embedded depth},
ylabel={error rate},
xmin=-0.15, xmax=23,
ymin=-0.0249906967827783, ymax=0.54713235377034,
width=\figurewidth,
height=\figureheight,
ytick={0,0.1,0.2,0.3,0.4,0.5},
tick align=outside,
tick pos=left,
x grid style={lightgray!92.02614379084967!black},
y grid style={lightgray!92.02614379084967!black}
]
\addplot [very thick, color0, mark=x, mark size=3, mark options={solid}, forget plot]
table {%
2 0.0395812349296463
4 0.126885914766741
6 0.255980653130818
8 0.280518837120274
10 0.321648199053476
12 0.385987261146497
14 0.446351691361476
16 0.477889400056333
18 0.506163615988046
20 0.489519112207152
22 0.504009163802978
24 0.52112676056338
26 0.440860215053763
28 0.64
30 0.4
32 0
};
\addplot [very thick, color1, mark=x, mark size=3, mark options={solid}, forget plot]
table {%
1 0.00101489642418162
3 0.00297947291313438
5 0.00421100551983156
7 0.0159726636999364
9 0.0369828456104945
11 0.0776109869051421
13 0.145334146553977
15 0.219062368013515
17 0.319633507853403
19 0.397016637980493
21 0.466045272969374
23 0.48
25 0.614285714285714
27 0.545454545454545
29 0.5
39 1
};
\draw[] (axis cs:12,0.446351691361476) -- (axis cs:12,0.446351691361476);
\node at (axis cs:12,0.416351691361476)[
  anchor=south east,
  text=color0,
  rotate=0.0
]{ out-of-sample};
\draw[] (axis cs:15,0.135334146553977) -- (axis cs:15,0.135334146553977);
\node at (axis cs:15,0.165334146553977)[
  anchor=north west,
  text=color1,
  rotate=0.0
]{ in-sample};
\end{axis}

\end{tikzpicture}

%% file: figures/generalization/distance_random.tex
% This file was created by matplotlib2tikz v0.6.17.
\begin{tikzpicture}

\definecolor{color1}{rgb}{0.83921568627451,0.152941176470588,0.156862745098039}
\definecolor{color0}{rgb}{0.12156862745098,0.466666666666667,0.705882352941177}

\begin{axis}[
xlabel={distance},
ylabel={error rate},
xmin=-3.9, xmax=103.9,
ymin=-0.0235290868119367, ymax=0.548170652110499,
width=\figurewidth,
height=\figureheight,
tick align=outside,
tick pos=left,
ytick = {0,0.1,0.2,0.3,0.4,0.5},
x grid style={lightgray!92.02614379084967!black},
y grid style={lightgray!92.02614379084967!black}
]
\addplot [very thick, color0, mark=x, mark size=3, mark options={solid}, forget plot]
table {%
1 0.158144068956082
3 0.123450541562756
7 0.114226519337017
13 0.14181630613495
17 0.178298397040691
21 0.215888646908437
23 0.234215681350143
27 0.266901853411963
33 0.324035332403533
35 0.336630659060566
41 0.381540549711571
45 0.399160625715376
49 0.412888850801268
51 0.423256649892164
53 0.420822646527976
55 0.424385823600483
57 0.425559803386128
65 0.441343884492573
67 0.446840597194993
69 0.445623342175066
71 0.444724462365591
77 0.455364134690681
79 0.465689398767793
93 0.503521126760563
99 0.522184300341297
};
\addplot [very thick, color1, mark=x, mark size=3, mark options={solid}, forget plot]
table {%
5 0.00311911942876297
9 0.00264224062004581
11 0.00245726495726495
15 0.00266036062666275
19 0.00495926319518247
25 0.0127646326276464
29 0.0227998678268532
31 0.0304156809733018
37 0.0566523605150214
39 0.0572460030943786
43 0.0743742889647326
47 0.098501753267453
59 0.141368373357499
61 0.154634146341463
63 0.155816435432231
73 0.20014245014245
75 0.194928084784254
81 0.225747960108794
83 0.230994152046784
85 0.221518987341772
87 0.245777518928363
89 0.229357798165138
91 0.286439448875997
95 0.285553047404063
97 0.301164725457571
};
\draw[] (axis cs:49,0.421184919210054) -- (axis cs:49,0.421184919210054);
\node at (axis cs:49,0.401184919210054)[
  scale=1,
  anchor=south east,
  text=color0,
  rotate=0.0
]{ out-of-sample};
\draw[] (axis cs:51,0.0893746831164442) -- (axis cs:51,0.0893746831164442);
\node at (axis cs:55,0.0993746831164442)[
  scale=1,
  anchor=north west,
  text=color1,
  rotate=0.0
]{ in-sample};
\end{axis}

\end{tikzpicture}

%% file: figures/generalization/depth_increase_random.tex
% This file was created by matplotlib2tikz v0.6.17.
\begin{tikzpicture}

\definecolor{color1}{rgb}{0.83921568627451,0.152941176470588,0.156862745098039}
\definecolor{color0}{rgb}{0.12156862745098,0.466666666666667,0.705882352941177}

\begin{axis}[
xlabel={embedded depth},
ylabel={error rate},
xmin=-0.9, xmax=24,
ymin=-0.05, ymax=0.55,
width=\figurewidth,
height=\figureheight,
tick align=outside,
tick pos=left,
ytick={0,0.1,0.2,0.3,0.4,0.5},
x grid style={lightgray!92.02614379084967!black},
y grid style={lightgray!92.02614379084967!black}
]
\addplot [very thick, color0, mark=x, mark size=3, mark options={solid}, forget plot]
table {%
3 0.00650710731759818
6 0.199634496562527
7 0.252274974274127
8 0.266357454856577
13 0.450149041242159
20 0.488036303630363
21 0.511186440677966
22 0.501149425287356
23 0.487755102040816
25 0.45578231292517
26 0.591397849462366
29 0.3
30 0.8
31 1
32 1
39 0
};
\addplot [semithick, color1, mark=x, mark size=3, mark options={solid}, forget plot]
table {%
1 0.000192164221282765
2 0.000362783273272549
4 0.00190992271475532
5 0.00691257295030878
9 0.166928177354862
10 0.187202063821435
11 0.197541870980067
12 0.211313490300607
14 0.245065789473684
15 0.270534095663087
16 0.288748137108793
17 0.341540073336826
18 0.343587842846553
19 0.357347055460263
24 0.431654676258993
27 0.565217391304348
28 0.5
};
\draw[] (axis cs:12,0.446351691361476) -- (axis cs:12,0.446351691361476);
\node at (axis cs:10,0.366351691361476)[
  anchor=south east,
  text=color0,
  rotate=0.0
]{ out-of-sample};
\draw[] (axis cs:15,0.135334146553977) -- (axis cs:15,0.135334146553977);
\node at (axis cs:15,0.25334146553977)[
  anchor=north west,
  text=color1,
  rotate=0.0
]{ in-sample};
\end{axis}

\end{tikzpicture}

%% file: figures/generalization/distance_extrapolation.tex
% This file was created by matplotlib2tikz v0.6.17.
\begin{tikzpicture}

\definecolor{color1}{rgb}{0.83921568627451,0.152941176470588,0.156862745098039}
\definecolor{color0}{rgb}{0.12156862745098,0.466666666666667,0.705882352941177}

\begin{axis}[
xlabel={distance},
ylabel={error rate},
xmin=-3.9, xmax=50,
ymin=-0.0264248704663212, ymax=0.554922279792746,
width=\figurewidth,
height=\figureheight,
tick align=outside,
tick pos=left,
ytick={0,0.1,0.2,0.3,0.4,0.5},
x grid style={lightgray!92.02614379084967!black},
y grid style={lightgray!92.02614379084967!black}
]
\addplot [thick, black, dashed, forget plot]
table {%
12 -1
12 1
};
\addplot [very thick, color0, forget plot]
table {%
13 0.0409841351734812
15 0.130117551682205
17 0.241901003124486
19 0.336448598130841
21 0.414574054771155
23 0.445002259149556
25 0.484865850481896
27 0.489423279309619
29 0.496821679491469
31 0.500594126633848
33 0.503533364226135
35 0.503982973548191
37 0.497455870343615
39 0.505169480434498
41 0.494701086956522
43 0.494252037252619
45 0.504814305364512
47 0.491463220041487
49 0.495675999657505
51 0.496269662921348
53 0.503946742748455
55 0.503022365504735
57 0.491874795506598
59 0.492406810860561
61 0.512741217219198
63 0.499603698811096
65 0.503122831367106
67 0.502940732921128
69 0.50054661877245
71 0.491334342924449
73 0.49723960250276
75 0.509329779131759
77 0.503035049931467
79 0.491190829972405
81 0.49930426716141
83 0.4957451981522
85 0.497744759883258
87 0.507502206531333
89 0.496551724137931
91 0.501669758812616
93 0.505272407732865
95 0.502290950744559
97 0.506415739948674
99 0.528497409326425
};
\addplot [very thick, color1, forget plot]
table {%
1 0
3 0.000127474368546654
5 2.57300913417691e-05
7 0.000103096326334207
9 8.7581012436555e-05
11 0.00285960936656959
};
\draw[] (axis cs:55,0.491874795506598) -- (axis cs:55,0.491874795506598);
\node at (axis cs:16,0.2)[
  anchor=north west,
  text=color0,
  rotate=0.0
]{ out-of-sample};
\draw[] (axis cs:9,-0.00989690367366579) -- (axis cs:9,-0.00989690367366579);
\node at (axis cs:10.5,0.2)[
  anchor=north east,
  text=color1,
  rotate=0.0
]{ in-sample};
\end{axis}

\end{tikzpicture}

%% file: figures/generalization/depth_increase_extrapolate.tex
% This file was created by matplotlib2tikz v0.6.17.
\begin{tikzpicture}

\definecolor{color1}{rgb}{0.83921568627451,0.152941176470588,0.156862745098039}
\definecolor{color0}{rgb}{0.12156862745098,0.466666666666667,0.705882352941177}

\begin{axis}[
xlabel={embedded depth},
ylabel={error rate},
xmin=-0.9, xmax=28,
ymin=-0.05, ymax=0.55,
width=\figurewidth,
height=\figureheight,
tick align=outside,
tick pos=left,
ytick={0,0.1,0.2,0.3,0.4,0.5},
x grid style={lightgray!92.02614379084967!black},
y grid style={lightgray!92.02614379084967!black}
]
\addplot [thick, black, dashed, forget plot]
table {%
12.5 -1
12.5 1
};
\addplot [very thick, color0, forget plot]
table {%
13 0.427199573200551
14 0.465473860777188
15 0.486899098203508
16 0.506605453012274
17 0.505831476870659
18 0.504116766467066
19 0.504913294797688
20 0.511321531494442
21 0.50471063257066
22 0.495454545454545
23 0.503067484662577
24 0.5
25 0.486301369863014
26 0.483870967741935
27 0.478260869565217
28 0.44
29 0.5
30 0.2
31 0
32 1
39 1
};
\addplot [very thick, color1, forget plot]
table {%
1 0.00081232018954136
2 0.000351357572967226
3 0.000360791773947544
4 0.000378880741714771
5 0.000789510785281222
6 0.00193143408981167
7 0.00330901407327677
8 0.00606493043168033
9 0.00833712293466726
10 0.017767781838628
11 0.0277372262773723
12 0.0622501850481125
};
\node at (axis cs:14,0.45)[
  anchor=north west,
  text=color0,
  rotate=0.0
]{ out-of-sample};
\node at (axis cs:11,0.15)[
  anchor=north east,
  text=color1,
  rotate=0.0
]{ in-sample};
\end{axis}

\end{tikzpicture}

%% file: figures/analysis_depth_results.tex
% This file was created by matplotlib2tikz v0.6.17.
\begin{tikzpicture}

\definecolor{color1}{rgb}{1,0.498039215686275,0.0549019607843137}
\definecolor{color0}{rgb}{0.12156862745098,0.466666666666667,0.705882352941177}
\definecolor{color3}{rgb}{0.83921568627451,0.152941176470588,0.156862745098039}
\definecolor{color2}{rgb}{0.172549019607843,0.627450980392157,0.172549019607843}

\begin{axis}[
xlabel={depth},
ylabel={error},
xmin=-3.45, xmax=52,
ymin=-2.89140968826998, ymax=40,
width=\figurewidth,
height=\figureheight,
tick align=outside,
tick pos=left,
x grid style={lightgray!92.02614379084967!black},
y grid style={lightgray!92.02614379084967!black},
legend entries={{2 hidden units},{8 hidden units},{50 hidden units},{predicting mean depth}},
legend cell align={left},
legend style={at={(0.03,0.97)}, anchor=north west, draw=white!80.0!black}
]

\addplot [very thick, color0, mark=x, mark size=3, mark repeat=5, mark options={solid}]
table {%
0 5.68262519648131
1 4.95989248828865
2 6.09702061232199
3 7.24121994112566
4 7.5260685716609
5 7.23083108980984
6 6.5970362619627
7 5.87622129325157
8 5.13460314453172
9 4.44261917547585
10 3.82340160348228
11 3.30499613713464
12 2.9780506454174
13 2.78266433020382
14 2.81456694584099
15 3.06120690722888
16 3.54403262600844
17 4.17748973506125
18 4.92343571742336
19 5.79041268478581
20 6.71710554309564
21 7.6249485039386
22 8.5862425476733
23 9.62378891416703
24 10.6359089075398
25 11.5608409399653
26 12.4536560235038
27 13.3390529080524
28 14.3455934787718
29 15.3810587307053
30 16.3101662413867
31 17.315501968796
32 18.353530487867
33 19.435909190611
34 20.5742250908641
35 21.4796655919623
36 22.3578331965336
37 23.3239831935952
38 24.3840678196219
39 25.40660691134
40 26.2380298544927
41 27.1709751625101
42 28.2048870776203
43 29.1313707278663
44 30.3520825120964
45 31.3474098266191
46 32.1373290561982
47 33.4894289998075
48 34.5004431271317
49 35.0643250435214
50 36.0741871013198
51 37.2455176186973
52 37.8992708955107
53 39.1984378348152
54 40.2920343332123
55 40.8548612225916
56 41.9236630684621
57 42.8474043352263
58 43.9814671278
59 44.6388806012961
60 43.7216735415988
61 45.3304101626078
62 48.0397672653198
63 49.8646801312765
64 50.1289143562317
65 50.4471003214518
66 52.8526711463928
67 51.629506111145
68 51.8367678324381
69 52.7479763031006
};
\addplot [very thick, color1, mark=*, mark size=3, mark repeat=5, mark options={solid}]
table {%
0 0.554440841489666
1 0.841311107540141
2 0.97734448611786
3 1.27005156611693
4 1.52006141750997
5 1.79143916129116
6 2.12919068286518
7 2.51233308146542
8 2.93025725016152
9 3.35098193490517
10 3.77709763013227
11 4.16393283491603
12 4.41083483221777
13 4.5244335381775
14 4.53031212704622
15 4.42481434404406
16 4.28897080766489
17 4.13486453343042
18 4.06827321243817
19 4.0426935114397
20 4.14182226643746
21 4.32074279443947
22 4.46263912044536
23 4.80122136363576
24 5.21908060586089
25 5.69128689275077
26 6.18742203360746
27 6.85218682191461
28 7.57846856202773
29 8.3393827936056
30 9.22401537086577
31 10.1008185153454
32 10.9756415617963
33 11.8291599437692
34 12.6467969399955
35 13.5967524460321
36 14.5502247183338
37 15.5156635285383
38 16.39043645029
39 17.2722530563566
40 18.2103623457444
41 19.0485613056041
42 20.0006116736146
43 20.8161975686401
44 21.7201924470433
45 22.6250414696766
46 23.2921658394504
47 24.2234740597296
48 25.2902079213964
49 26.1019569348685
50 27.1143258623903
51 28.3812689164589
52 29.01847652931
53 29.7101402831592
54 30.9913820467497
55 32.0359897121941
56 32.8181062646814
57 34.3657804897853
58 35.9111771800301
59 37.853564445789
60 36.6267549726698
61 38.4359450340271
62 40.6605889002482
63 43.0636355082194
64 44.5145101547241
65 44.2312787373861
66 45.5504474639893
67 45.6235224405924
68 46.5634250640869
69 45.6111602783203
};
\addplot [very thick, color2, mark=triangle*, mark size=3, mark repeat=5, mark options={solid}]
table {%
0 0.0381670105202214
1 0.196681495188713
2 0.213331791104516
3 0.258769134025913
4 0.297410084072427
5 0.344894752523432
6 0.415474283909572
7 0.49339824672366
8 0.591032922007132
9 0.683570103495181
10 0.80606605330691
11 0.929465294426047
12 1.0840478144034
13 1.25285178251265
14 1.4285150811461
15 1.6119726248104
16 1.84893910248551
17 2.06756778649847
18 2.27135155144974
19 2.46556360016027
20 2.65254072552254
21 2.80182819404082
22 2.89623075279883
23 2.98456644028368
24 3.06615499243135
25 3.16174578182871
26 3.25319408969856
27 3.32947559808648
28 3.38398496847683
29 3.49663940462723
30 3.68617537934007
31 3.7990507851644
32 3.93051957482862
33 4.15934243587412
34 4.35885445394344
35 4.67906315247291
36 5.08139457953778
37 5.47838847469276
38 6.08871906041546
39 6.63483727379273
40 7.00418421171233
41 7.76560345782509
42 8.51433664205186
43 8.8846396256197
44 9.35818613412087
45 10.191351848593
46 10.8442229597013
47 11.865469336312
48 12.7478133211041
49 13.3650127410889
50 14.2607405779766
51 15.2547666204387
52 16.5058667888749
53 17.6868674737944
54 18.5143198214079
55 18.8314618572746
56 19.3490833334021
57 20.2863483088357
58 20.9615781957453
59 22.240769973168
60 24.8209830390082
61 23.3778416315715
62 20.7143478393555
63 23.3158041636149
64 27.2666997909546
65 28.9259401957194
66 30.6595058441162
67 32.0451062520345
68 32.0385475158691
69 32.0849685668945
};
\addplot [very thick, color3, dashed]
table {%
0 11.3702990136757
1 9.9417275851043
2 8.51315615653287
3 7.08458472796144
4 5.65601329939001
5 4.22744187081859
6 2.79887044224716
7 1.37029901367573
8 0.0582724148957006
9 1.48684384346713
10 2.91541527203856
11 4.34398670060999
12 5.77255812918141
13 7.20112955775284
14 8.62970098632427
15 10.0582724148957
16 11.4868438434671
17 12.9154152720386
18 14.34398670061
19 15.7725581291814
20 17.2011295577528
21 18.6297009863243
22 20.0582724148957
23 21.4868438434671
24 22.9154152720386
25 24.34398670061
26 25.7725581291814
27 27.2011295577528
28 28.6297009863243
29 30.0582724148957
30 31.4868438434671
31 32.9154152720386
32 34.34398670061
33 35.7725581291814
34 37.2011295577528
35 38.6297009863243
36 40.0582724148957
37 41.4868438434671
38 42.9154152720386
39 44.34398670061
40 45.7725581291814
41 47.2011295577528
42 48.6297009863243
43 50.0582724148957
44 51.4868438434671
45 52.9154152720386
46 54.34398670061
47 55.7725581291814
48 57.2011295577528
49 58.6297009863243
};
\end{axis}

\end{tikzpicture}

%% file: figures/analysis_previous_results.tex
% This file was created by matplotlib2tikz v0.6.17.
\begin{tikzpicture}

\definecolor{color0}{rgb}{0.12156862745098,0.466666666666667,0.705882352941177}

\begin{axis}[
xlabel={k-to-last symbol},
ylabel={error rate},
xmin=-4.9, xmax=102.9,
ymin=-0.02519796145765, ymax=0.532570187761043,
width=\figurewidth,
height=\figureheight,
tick align=outside,
tick pos=left,
ytick = {0,0.1,0.2,0.3,0.4,0.5},
x grid style={lightgray!92.02614379084967!black},
y grid style={lightgray!92.02614379084967!black},
legend style={at={(0.97,0.03)}, anchor=south east, draw=white!80.0!black},
legend cell align={left},
legend entries={{relevant symbols},{irrelevant symbols}}
]
\addplot [very thick, black]
table {%
0 1.63301299062057e-05
1 0.000950950261166761
2 0.00197882792499826
3 0.0119352508536905
4 0.0177724922782155
5 0.0401395204441279
6 0.062843893192985
7 0.0863014238946082
8 0.0999801311345122
9 0.119206545165408
10 0.128620282023136
11 0.144828498368751
12 0.156037991858887
13 0.167759058389107
14 0.176038777935727
15 0.185634507991527
16 0.193106130584861
17 0.201222720553231
18 0.205968788001621
19 0.212328231336862
20 0.217065710368105
21 0.224858020162496
22 0.230604218805863
23 0.23745828855834
24 0.238020221693925
25 0.24361088211047
26 0.248051133478689
27 0.253894128851885
28 0.260826219702162
29 0.266267643952245
30 0.268953185955787
31 0.273252899301494
32 0.277847564080801
33 0.283654517525485
34 0.289581187169195
35 0.294283002336449
36 0.300375143928983
37 0.305097050282745
38 0.308970616851527
39 0.313565098841172
40 0.319560241948668
41 0.328127630914295
42 0.331671894469987
43 0.339265161575919
44 0.338190704620537
45 0.344343595021989
46 0.345669555225739
47 0.35315654066891
48 0.357764729702369
49 0.357811111691841
50 0.364723548023811
51 0.37119528993557
52 0.368832354115567
53 0.374489976938088
54 0.375958145759825
55 0.381193124368049
56 0.380744410403494
57 0.387391657638137
58 0.390759537976899
59 0.40064219513975
60 0.404097678989945
61 0.413923750887994
62 0.411195803622654
63 0.423696844993141
64 0.420789872514933
65 0.423661971830986
66 0.420830470207127
67 0.42205441191683
68 0.422159583694709
69 0.429621125143513
70 0.425827494563904
71 0.435431955710055
72 0.431626259874693
73 0.437318630295995
74 0.437413607740746
75 0.448571900280667
76 0.446210916799152
77 0.440923840427562
78 0.442918630866981
79 0.446998437848695
80 0.448954788526981
81 0.450147019513499
82 0.456642335766423
83 0.46133853151397
84 0.476415094339623
85 0.473100616016427
86 0.468981481481481
87 0.476012793176972
88 0.480639213275968
89 0.4745269286754
90 0.471204188481675
91 0.476973684210526
92 0.495901639344262
93 0.505208333333333
94 0.510250569476082
95 0.459183673469388
96 0.452631578947368
97 0.423913043478261
98 0.53125
};
\addplot [very thick, black, dashed]
table {%
0 0.00203919653214479
1 0.00922295802311945
2 0.10096508409792
3 0.22110250040905
4 0.331485049380281
5 0.396054459159344
6 0.431628202798474
7 0.449432699986439
8 0.458954048509899
9 0.464106535962823
10 0.466423569907346
11 0.470303730734827
12 0.47302961533181
13 0.472939448343398
14 0.475729206576602
15 0.475882779680462
16 0.479791192888154
17 0.480269767360595
18 0.482569396732487
19 0.481684168561824
20 0.481927710843373
21 0.481525726648481
22 0.481682039838798
23 0.481444519455615
24 0.483260436579824
25 0.485112193531294
26 0.48460423552965
27 0.484947630074996
28 0.486552061799587
29 0.485128296191068
30 0.486156749934621
31 0.486595952093585
32 0.488272743499732
33 0.487425091470456
34 0.488117799610992
35 0.489314495562583
36 0.489656165710844
37 0.487282570823618
38 0.48907190370605
39 0.49110731865559
40 0.493231970055276
41 0.491690348306081
42 0.493290441176471
43 0.492283489341619
44 0.494528121402746
45 0.494229520209169
46 0.495678322879972
47 0.4964544023767
48 0.4970938733733
49 0.495628085779972
50 0.495176096828893
51 0.495948038367954
52 0.492801333472237
53 0.493247172859451
54 0.495227663902363
55 0.492965893840529
56 0.49133225938304
57 0.492701549517179
58 0.493374902572097
59 0.495111279170268
60 0.497307811575001
61 0.496856085091379
62 0.49395886494297
63 0.495145006751977
64 0.495914094684946
65 0.496268524740919
66 0.495735409140005
67 0.496471418489767
68 0.49744993158353
69 0.49553101997897
70 0.498067700330586
71 0.498270921845667
72 0.494724068798143
73 0.493088857545839
74 0.492621606850064
75 0.491970231100666
76 0.497773694254011
77 0.494439757650127
78 0.49685534591195
79 0.501376146788991
80 0.499747347145023
81 0.498826946709865
82 0.502937132858393
83 0.508332166363255
84 0.508719086873811
85 0.508483754512635
86 0.514375
87 0.505828071879553
88 0.507794457274827
89 0.510726643598616
90 0.506773920406435
91 0.50920568122041
92 0.498641304347826
93 0.4967978042086
94 0.49738219895288
95 0.485380116959064
96 0.456591639871383
97 0.409937888198758
98 0.392857142857143
};
\end{axis}

\end{tikzpicture}